\documentclass{article}

\usepackage{arxiv}

\usepackage[utf8]{inputenc} 
\usepackage[T1]{fontenc}    
\usepackage{hyperref}       
\usepackage{url}            
\usepackage{booktabs}       
\usepackage{amsfonts}       
\usepackage{nicefrac}       
\usepackage{microtype}      
\usepackage{lipsum}
\usepackage{graphicx}
\usepackage[caption=false,font=scriptsize ,labelfon
t=sf,textfont=sf]{subfig}

\title{Classification Performance Metric for Imbalance Data Based on Recall and Selectivity Normalized in Class Labels}

\author{
  Robert Burduk  \\
  Wroclaw University of Science and Technology\\
  Wybrzeze Wyspianskiego 27\\
  50-370, Wroclaw, Poland \\
  \texttt{robert.burduk@pwr.edu.pl} \\
}

\begin{document}
\maketitle

\begin{abstract}
In the classification of a class imbalance dataset, the performance measure used for the model selection and comparison to competing methods is a major issue. In order to overcome this problem several performance measures are defined and analyzed in several perspectives regarding in particular the imbalance ratio. There is still no clear indication which metric is universal and can be used for any skewed data problem. In this paper we introduced a new performance measure based on the harmonic mean of Recall
and Selectivity normalized in class labels. This paper shows that the proposed performance measure has the right properties for the imbalanced dataset. In particular, in the space defined by the majority class examples and imbalance ratio it is less sensitive to changes in the majority class and more sensitive to changes in the minority class compared with other existing single-value performance measures. Additionally, the identity of the other performance measures has been proven analytically.
\end{abstract}

\keywords{Performance metrics \and Imbalanced datasets \and  Model Selection \and Imbalance ratio.}

\section{Introduction}
One of the important threads in machine learning and data mining is the evaluation of the algorithm or classifier performance. In the case of a binary classification, classifier assessment methods use the data contained in the confusion matrix. Many metrics are considered to choose the right machine learning method for a specific life science problem. In bioinformatics, such problems are for example: the prediction of tissue condition based on gene expression, genome annotation or the gene classification. The performance measures, that have the intuitive visual interpretation are analyzed in
~\cite{flach2016roc}, \cite{grau2015prroc}, \cite{saito2017precrec}. Single-value metrics are also considered from different points of views~\cite{baldi2000assessing}, \cite{chicco2020advantages} \cite{japkowicz2011evaluating}.

In a wide range of scientific areas, including the life sciences, there is a class imbalance problem~\cite{arun2020class}, \cite{johnson2019survey}. The dataset is imbalanced when a difference in the numbers of positive and negative instances is significant. In bioinformatics unequal class distributions arise naturally~\cite{pes2020learning}, \cite{wald2013hidden}. It is well-known that the most commonly used performance measure, the overall accuracy, could not be used to measure the true performance of an imbalanced dataset. For this reason the selection of an appropriate performance measure to skewed data is an important problem. The selection and analysis of properties of various performance measures is still a current research problem regarding the imbalance data. One of the problems considered is the impact of class imbalance on the classification performance measures~\cite{luque2019impact}. The same problem with the imbalance factor is also considered for inbalanced and streaming data~\cite{brzezinski2019dynamics}. The work~\cite{saito2015precision} presents another comprehensive study of the differences between various measures from several perspectives for imbalanced datasets. The problem of the classification difficulty and performance measures is considered in~\cite{zhang2016classification}.
The generic performance measure for multiclass problems is instead introduced in~\cite{kautz2017generic}.

This research is focused on the definition of a new classification performance metric which has the desired properties for imbalanced data. In particular, this applies to the fact that they are less sensitive to changes in the majority class and more sensitive to changes in the minority class compared with other existing single-value performance measures. The results presented in this article extend the  discussion on the use of classification performance measures~\cite{chicco2020advantages}, \cite{delgado2019cohen}, in particular for imbalanced data~\cite{luque2019impact}, \cite{mullick2020appropriateness}.

The goal of this paper is to analyze the behavior of performance measures in the context of comparing two algorithms for imbalanced dataset. What is important in our analysis is the answer to the question:
Which machine learning method is better for the classification of the majority or minority class? The absolute value of the two values of performance measures can be used for this purpose. In addition, our analysis concerns the comparison of the proposed in this paper performance measure with other metrics considered in the article.

The rest of the paper is organized as follows. Section 2 presents the definition of the existing performance measures based on the confusion matrix. A new performance measure is defined in Section 3. In Section 4 we demonstrate that the proposed measure can be successfully used for imbalanced datasets. In particular, compared to other metrics. The conclusions of presented results are addressed in Section 5.

\section{Performance metrics}

In binary classification when a classifier produces a real-valued output all possible combinations of actual and predicted class labels form a confusion matrix. The confusion matrix is $2\times2$ table containing a number of four types of an outcome:
\begin{itemize}
\item True Positive -- $TP$,
\item True Negative -- $TN$,
\item False Positive -- $FP$,
\item False Negative -- $FN$.
\vspace*{1pt}
\end{itemize}
$TP$  and $TN$ items denote the number of examples classified correctly by the classifier as positive and negative respectively, while $FN$ and $FP$ indicate the number of  misclassified positive and negative examples respectively.
Additionally, $TP+FN=P$, $TN+FP=N$, $TP+FP=\widehat{P}$,  $TN+FN=\widehat{N}$, and $P+N=M$, where $M$ is a number of the test dataset examples, $P$ is a number of the positive and $N$ is a number of the negative test dataset examples, $\widehat{P}$ and $\widehat{N}$ are the predicted positive and negative number of  examples respectively.

In this work, we consider the following classification performance metrics which are the functions based on values from the confusion matrix:

\begin{equation}
\emph{Recall } (REC) =\frac{TP}{P}
\label{eq_REC}
\end{equation}

\begin{equation}
\emph{Precision } (PRC) =\frac{TP}{\widehat{P}}
\end{equation}

\begin{equation}
\emph{Selectivity } (SEL) =\frac{TN}{N}
\end{equation}

\begin{equation}
\emph{Accuracy } (ACC)=\frac{TP+TN}{M}
\label{eq_ACC}
\end{equation}
\begin{equation}
F_1-score =2\frac{PRC*REC}{PRC+REC}
\end{equation}
\begin{equation}
G-mean =\sqrt{REC*SEL}
\label{eq_GM}
\end{equation}
\begin{equation}
\emph{Matthews cor. coef. } (MCC) =\frac{TP*TN-FP*FN}{\sqrt{\widehat{P}*P*N*\widehat{N}}}
\label{eq_MCC}
\end{equation}
\begin{equation}
\emph{Kappa stat. } (Kappa) =\frac{ACC-\frac{1}{M}\left(\frac{P*\widehat{P}}{M}+ \frac{N*\widehat{N}}{M}\right)}{1-\frac{1}{M}\left(\frac{P*\widehat{P}}{M}+ \frac{N*\widehat{N}}{M}\right)}
\label{eq_Kappa}
\end{equation}

The Receiver Operating Characteristic ($ROC$) analysis is widely used for classifiers with the soft type of output. As a performance measure the Area Under Curve ($AUC$) is commonly used. For the real-valued output of the classifier $AUC$ and $BACC$ are equal~\cite{SOKOLOVA2009427}. Therefore, in this paper, whenever $BACC$ is mentioned it could be also replaced by $AUC$.

\begin{figure}[hbt!]
\centering
\subfloat[HMNC]{\includegraphics[width = 2in]{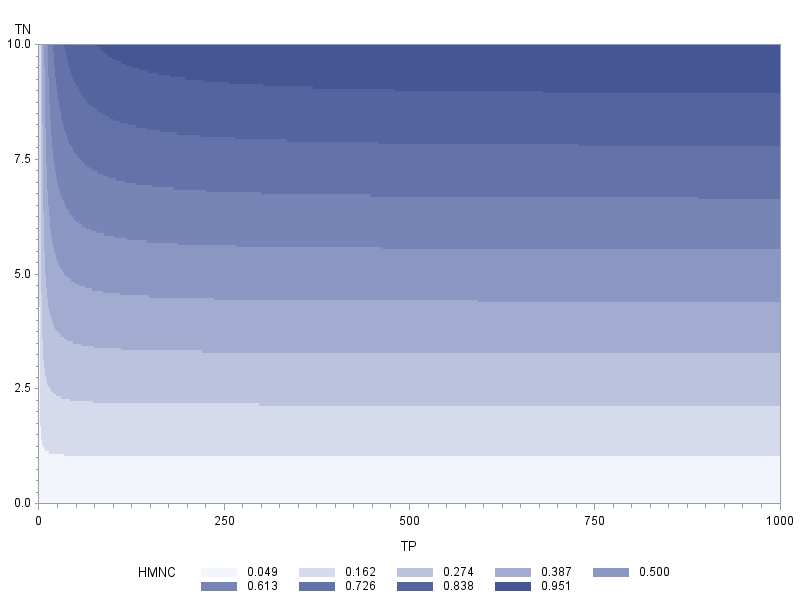}}
\subfloat[ACC]{\includegraphics[width = 2in]{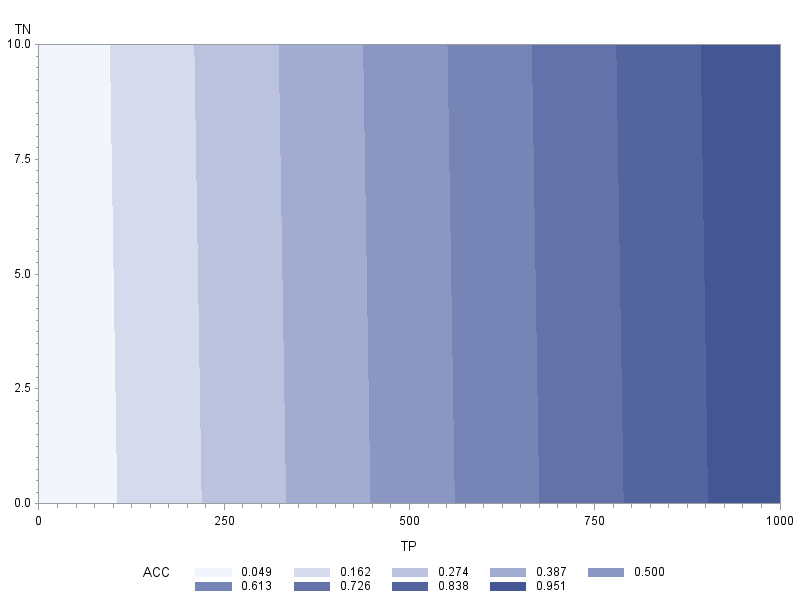}}\\
\subfloat[BACC]{\includegraphics[width = 2in]{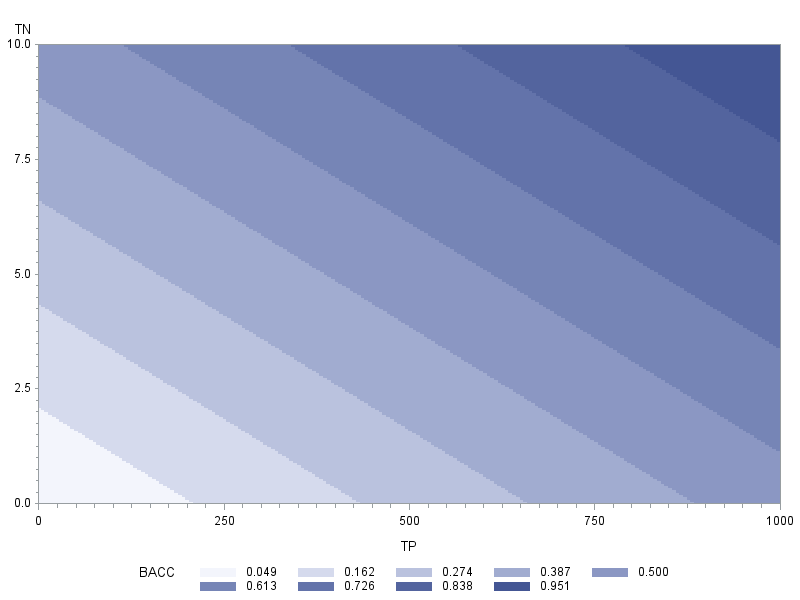}}
\subfloat[$F_1$-score]{\includegraphics[width = 2in]{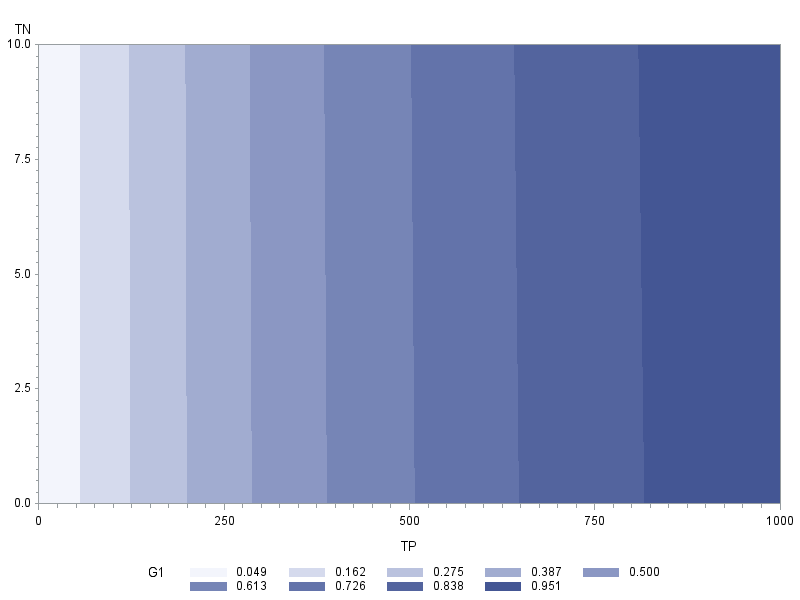}} \\
\subfloat[G-mean]{\includegraphics[width = 2in]{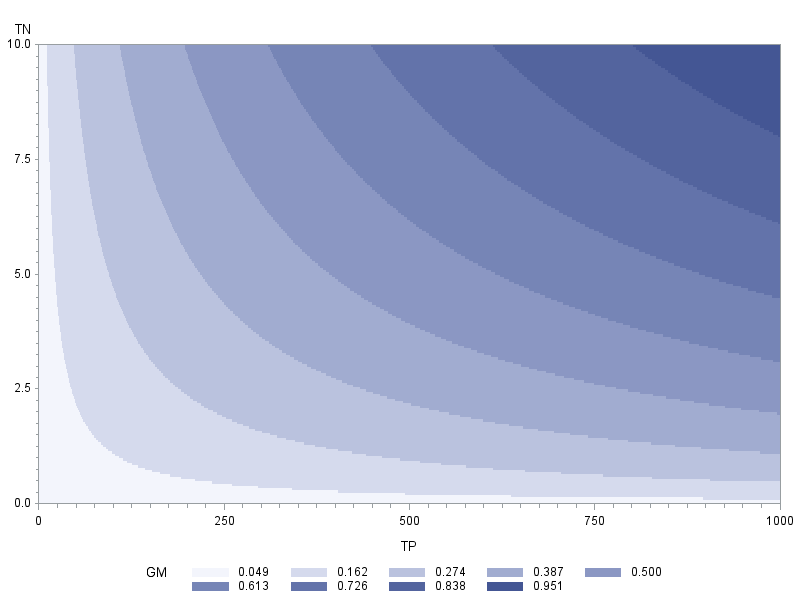}}
\subfloat[Kappa]{\includegraphics[width = 2in]{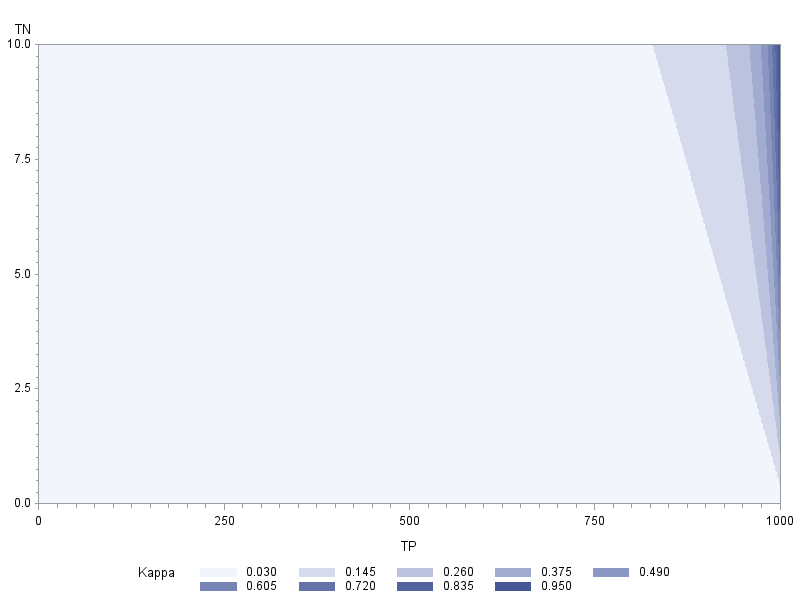}}\\
\subfloat[MCC]{\includegraphics[width = 2in]{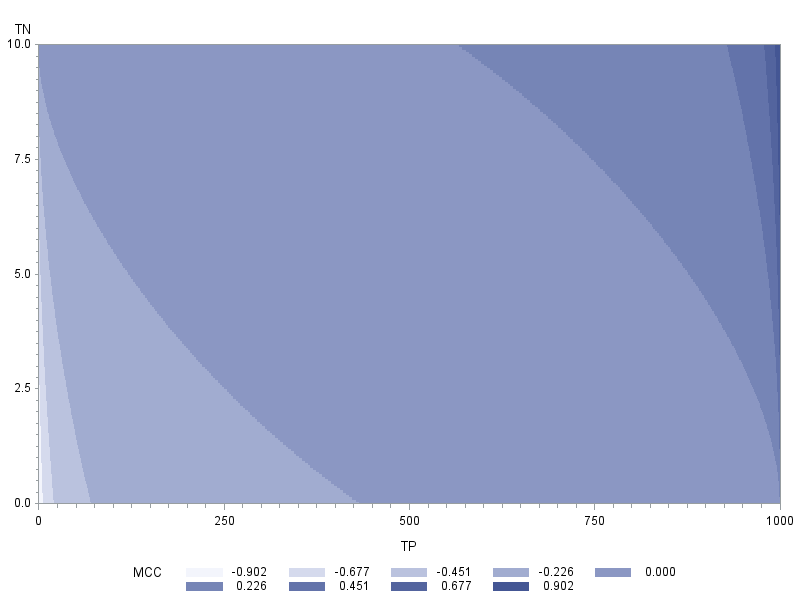}}

\caption{Set of heat maps for $IR=0.01$.}
\label{RB:Rys_IR001}
\end{figure}
\subsection{Imbalance ratio}

The class imbalance occurs when the difference between the number of positive and negative examples is high. In several articles there have been proposed different definitions of class imbalance~\cite{jeni2013facing}, \cite{luque2019impact}, \cite{zhu2020adjusting}.
In this paper we propose the imbalance ratio (IR) as
\begin{equation}
IR=\frac{\min(P,N)}{\max(P,N)}.
\label{eq:IR}
\end{equation}
This IR is independent of the fact which class (positive or negative) is the majority or minority class.

\section{A proposal of a new performance metric}

Our motivation is to use the measures: $REC$ and $SEL$ taking into account the normalization of class labels. The normalization refers to the number of class labels ($P$, $N$) in the number of all the objects $M$ from the test dataset. The normalized in class labels performance metrics are therefore the following: $REC*\frac{P}{M}$ and $SEL*\frac{N}{M}$. Given the above, we propose the following harmonic mean of $REC$
and $SEL$ normalized in the class labels ($HMNC$) performance metric:
\begin{equation}
HMNC=\frac{HM(REC*\frac{P}{M},SEL*\frac{N}{M})}{HM(\frac{P}{M},\frac{N}{M})}
\label{eq:HMN}
\end{equation}
where $HM$ is the harmonic mean.

Given the expression of the harmonic mean of two numbers $HM(a,b)=\frac{2ab}{a+b}$ the proposed performance metric can be expressed as:
\begin{equation}
HMNC=\frac{TP*TN*M}{(TP+TN)*P*N}.
\label{eq:HMNC}
\end{equation}
The domain of HMNC measure is between 0 and 1, where 1 is the preferred value which means that all objects are classified correctly. If the value is 0, each object is incorrectly classified. The same properties have performance measures~(\ref{eq_ACC})-(\ref{eq_GM}) while the measures~(\ref{eq_MCC}) and (\ref{eq_Kappa}) return values between $-1$ and $+1$.

The proposed metric has also a property, which is common for the other metrics as shown below.

\noindent
{\bf Theorem} {\it Performance measures $HMNC$, $ACC$, $BACC$ \emph{and} $G-mean$ \emph{are equal if} \\ $\frac{TP}{P}=\frac{TN}{N}$}.

\noindent
{\bf Proof}.
If the equation $\frac{TP}{P}=\frac{TN}{N}$ occurs, then equations $\frac{TP}{TN}=\frac{P}{N}$ and $TP*N=TN*P$ are true.

\noindent
Therefore:

 \setlength{\arraycolsep}{0.0em}
\begin{eqnarray}
HMNC&{}={}&\frac{TP*TN*M}{(TP+TN)*P*N}\nonumber\\
&{}={}&\frac{TP*TN*M}{TP*P*N+TP*N*N}\nonumber\\
&{}={}&\frac{TP*TN*M}{TP*N(P+N)}= \frac{TN}{N}\nonumber
\end{eqnarray}
\setlength{\arraycolsep}{5pt}

\setlength{\arraycolsep}{0.0em}
\begin{eqnarray}
ACC&{}={}&\frac{TP+TN}{M} = \frac{\frac{TN*P}{N}+TN}{P+N}= \frac{TN\left(\frac{P}{N}+1\right)}{P+N}\nonumber\\
&{}={}&\frac{TN\left(\frac{P+N}{N}\right)}{P+N}
= \frac{TN}{N}\nonumber
\end{eqnarray}
\setlength{\arraycolsep}{5pt}

\setlength{\arraycolsep}{0.0em}
\begin{eqnarray}
BACC&{}={}&0.5*\left(\frac{TP}{P}+\frac{TN}{N} \right)= 0.5*\left(\frac{TN}{N}+\frac{TN}{N} \right) =\frac{TN}{N} \nonumber
\end{eqnarray}
\setlength{\arraycolsep}{5pt}

\setlength{\arraycolsep}{0.0em}
\begin{eqnarray}
G-mean&{}={}&\sqrt{\frac{TP}{P}\frac{TN}{N}} = \sqrt{\frac{TN}{N}\frac{TN}{N}}= \frac{TN}{N} \nonumber
\end{eqnarray}
\setlength{\arraycolsep}{5pt}

\noindent
In the case of fulfilled assumptions of the theorem all performance measures $HMNC$, $ACC$, $BACC$ \emph{and} $G-mean$ are equal to $\frac{TN}{N}$ (or equivalent $\frac{TP}{P}$) which completes the proof.

\section{Analysis of HMNC performance metric}

The measures defined by equations~(\ref{eq_REC})-(\ref{eq_Kappa}) were discussed and compared in various articles~\cite{baldi2000assessing}, \cite{brzezinski2019dynamics}, \cite{SOKOLOVA2009427}. Additionally, visual-based analyses of performance measures are also presented, in particular ROC~\cite{flach2016roc} and precision-recall curves~\cite{saito2017precrec} are analyzed.

\begin{figure}[hbt!]
\centering
\subfloat[HMNC]{\includegraphics[width = 2in]{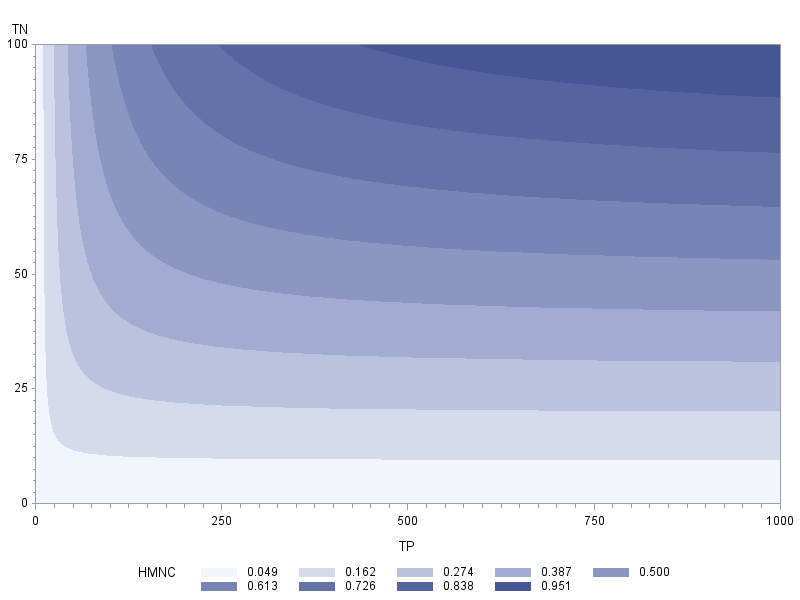}}
\subfloat[ACC]{\includegraphics[width = 2in]{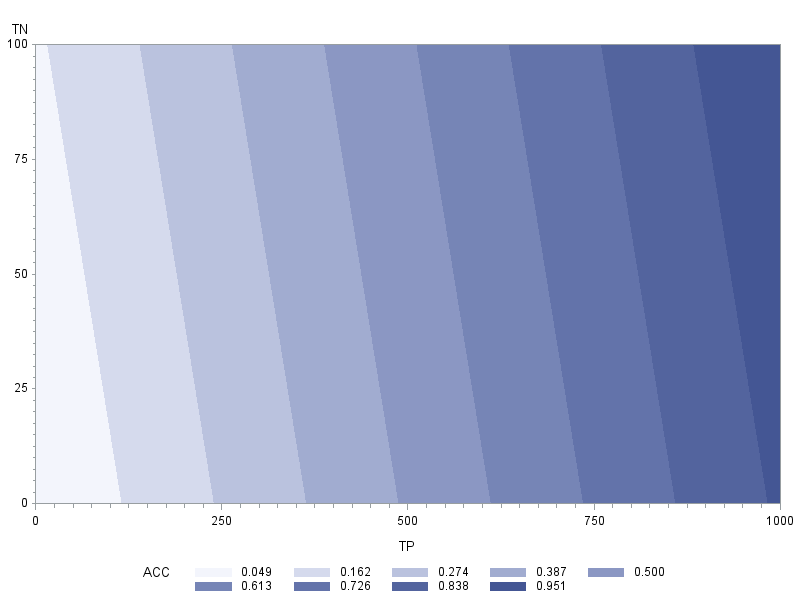}}\\
\subfloat[BACC]{\includegraphics[width = 2in]{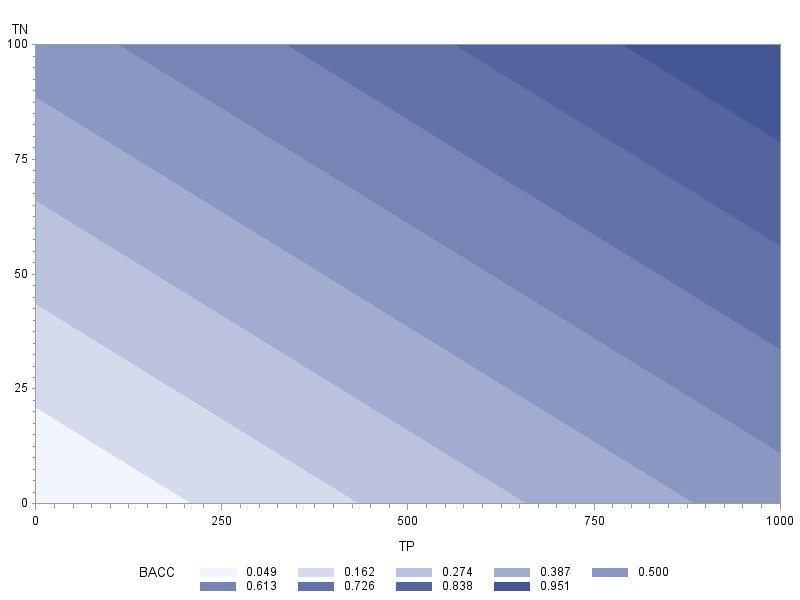}}
\subfloat[$F_1$-score]{\includegraphics[width = 2in]{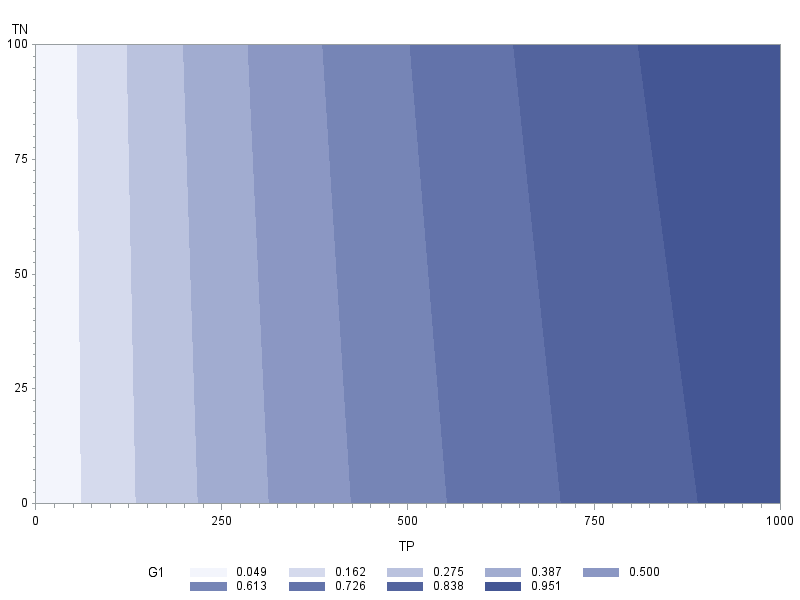}} \\
\subfloat[G-mean]{\includegraphics[width = 2in]{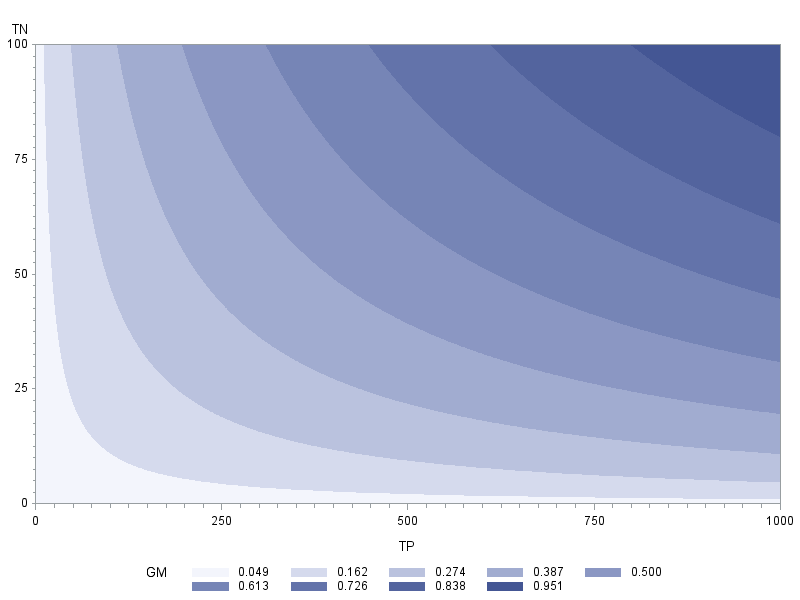}}
\subfloat[Kappa]{\includegraphics[width = 2in]{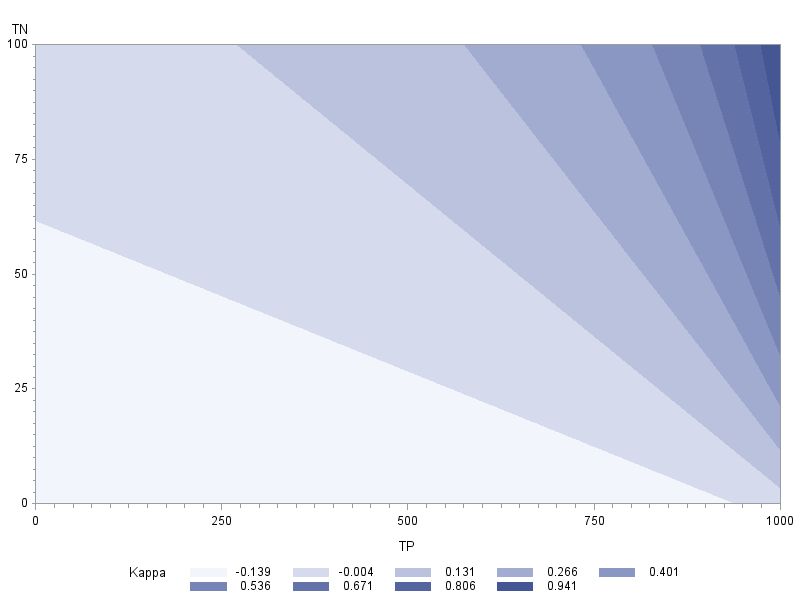}}\\
\subfloat[MCC]{\includegraphics[width = 2in]{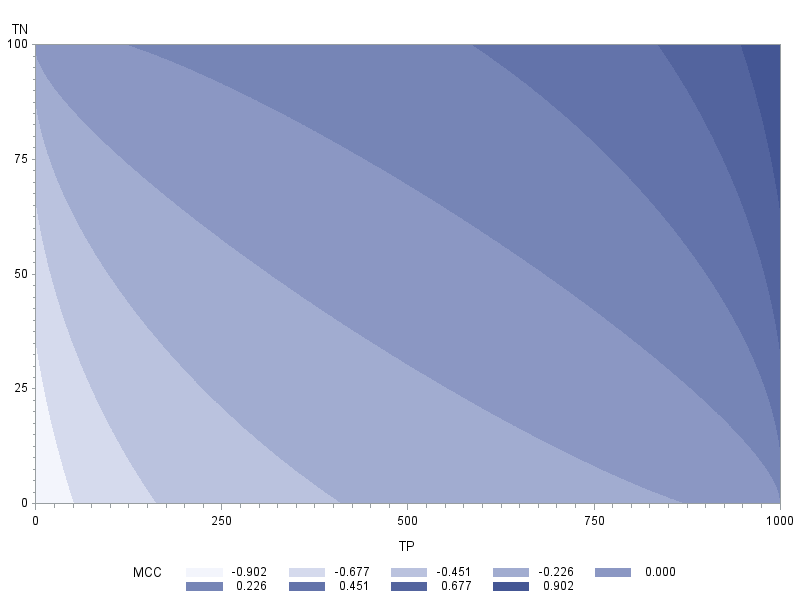}}

\caption{Set of heat maps for $IR=0.1$.}
\label{RB:Rys_IR01}
\end{figure}

\begin{figure}[hbt!]
\centering
\subfloat[HMNC]{\includegraphics[width = 2in]{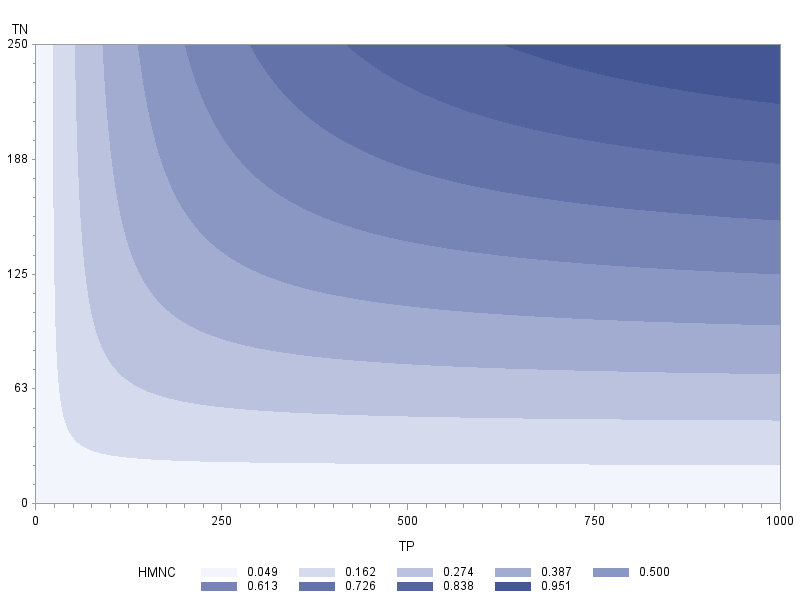}}
\subfloat[ACC]{\includegraphics[width = 2in]{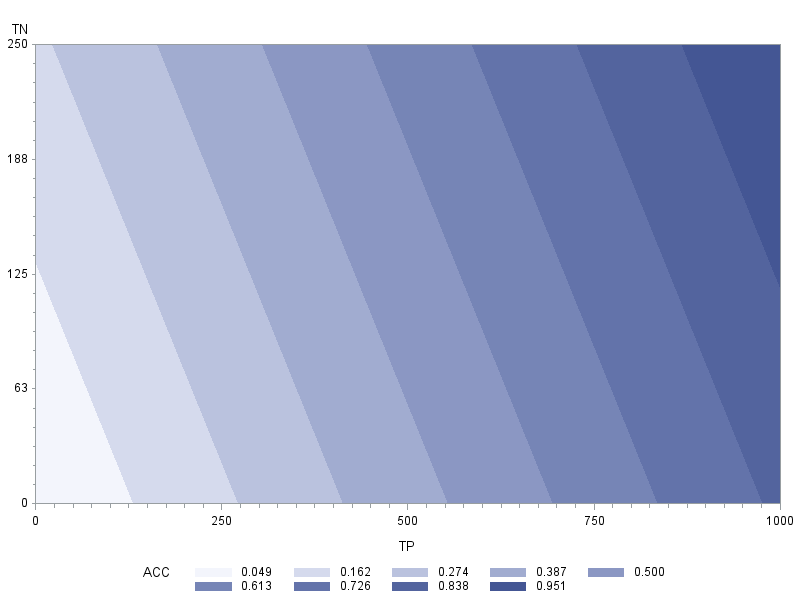}}\\
\subfloat[BACC]{\includegraphics[width = 2in]{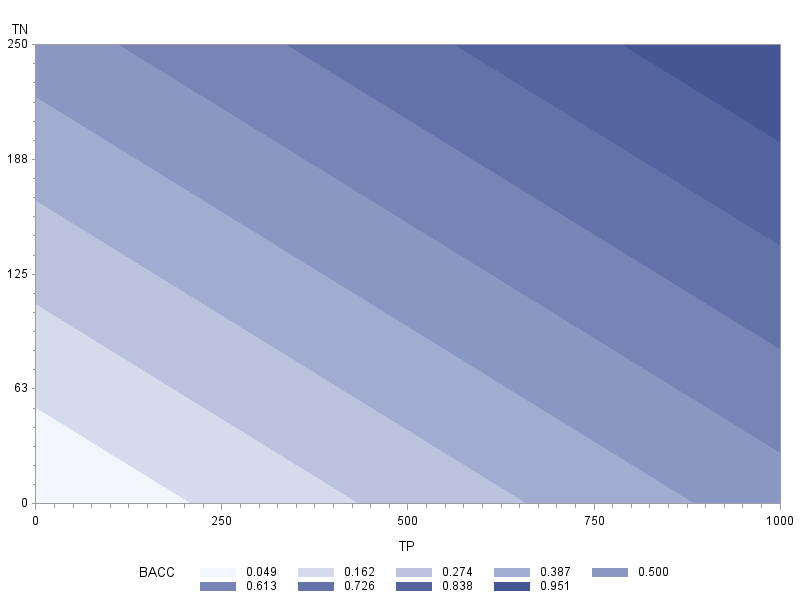}}
\subfloat[$F_1$-score]{\includegraphics[width = 2in]{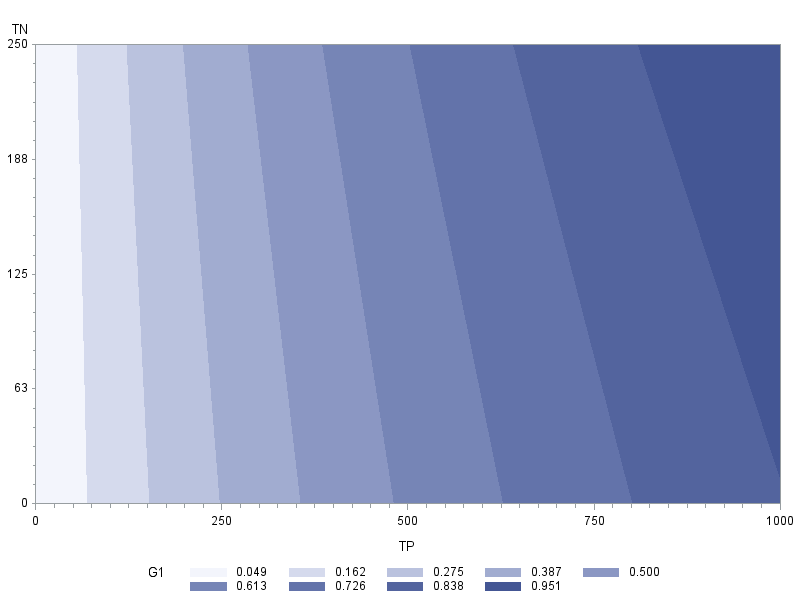}} \\
\subfloat[G-mean]{\includegraphics[width = 2in]{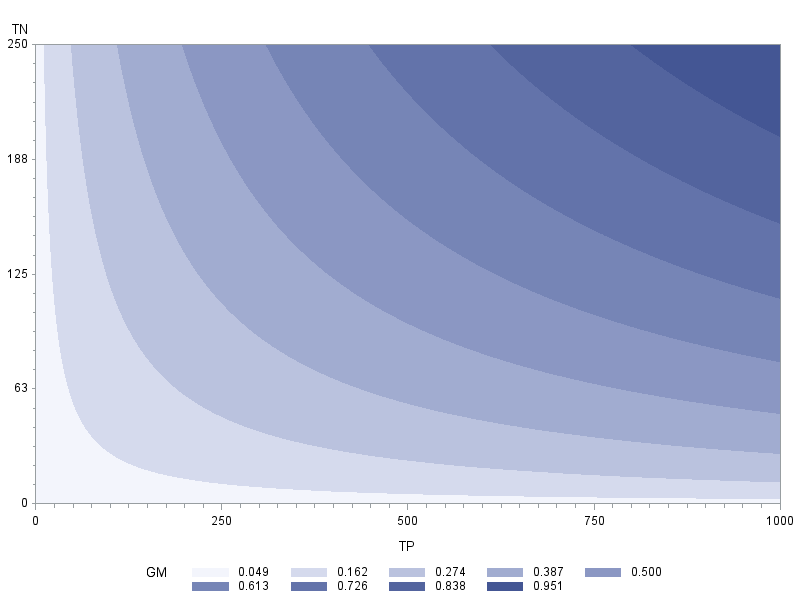}}
\subfloat[Kappa]{\includegraphics[width = 2in]{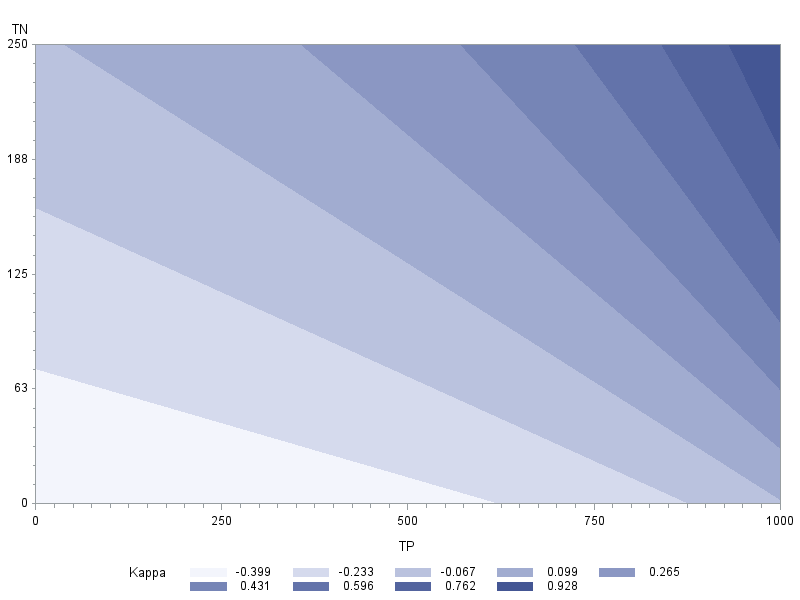}}\\
\subfloat[MCC]{\includegraphics[width = 2in]{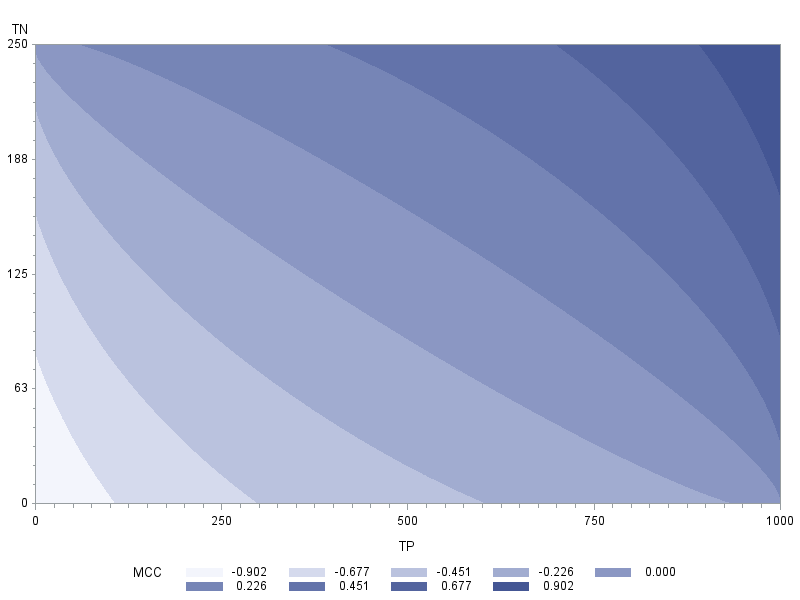}}

\caption{Set of heat maps for $IR=0.25$.}
\label{RB:Rys_IR025}
\end{figure}

\begin{table*}[hbt!]
\centering
\small
\caption{Values of performance metrics for four classifiers and theirs absolute value of the difference in measures -- the case of IR=0.01, P=1000, N=10.} \label{Tab:IR01}
\begin{tabular}{|c|c|c|c|c|c|c|c|c|c|}
\hline
Meth. & TP  & TN  & HMNC& ACC& BACC & MCC & $F_1$& $G-m$  & Kappa    \\
\hline
1      & 500 & 5 & 0.5  & 0.5  & 0.5  & 0    & 0.66 & 0.5  & 0    \\
2      & 700 & 5 & 0.5 & 0.7 & 0.6 & 0.04 & 0.82 & 0.59 & 0.01 \\
3      & 700 & 7 & 0.7 & 0.7 & 0.7 & 0.09 & 0.82 & 0.7 & 0.03 \\
4      & 500 & 7 & 0.7 & 0.5 & 0.6  & 0.04 & 0.67 & 0.59 & 0.01 \\\hline
\multicolumn{3}{|c|}{}     & \multicolumn{7}{c|}{Absolute value of the difference in measures}   \\ \hline
\multicolumn{3}{|c|}{$|\Psi_1-\Psi_2|$}  & 0.0 & 0.2 & 0.1 & 0.02 & 0.16 & 0.09 & 0.01 \\ \hline
\multicolumn{3}{|c|}{$|\Psi_1-\Psi_3|$}  & 0.2 & 0.2 & 0.2 & 0.04 & 0.16 & 0.09 & 0.01  \\ \hline
\multicolumn{3}{|c|}{$|\Psi_1-\Psi_4|$}  & 0.2 & 0.0 & 0.1  & 0.02 & 0.0 & 0.09 & 0.0 \\ \hline
\multicolumn{3}{|c|}{$|\Psi_2-\Psi_3|$} & 0.2 & 0.0 & 0.1  & 0.02 & 0.0 & 0.11  & 0.01 \\ \hline
\multicolumn{3}{|c|}{$|\Psi_3-\Psi_4|$} & 0.0 & 0.2 & 0.1 & 0.02 & 0.16 & 0.11 & 0.01 \\ \hline
\end{tabular}

\end{table*}

\begin{table*}[hbt!]
\centering
\small
\caption{Values of performance metrics for four classifiers and theirs absolute value of the difference in measures -- the case of IR=0.1, P=1000, N=100.} \label{Tab:IR1}
\begin{tabular}{|c|c|c|c|c|c|c|c|c|c|}
\hline
Meth. & TP  & TN  & HMNC       & ACC      & BACC & MCC      & $F_1$         & $G-m$  & Kappa    \\
\hline
1    & 500 & 50 & 0.5  & 0.5  & 0.5 & 0    & 0.65 & 0.5  & 0    \\ \hline
2    & 700 & 50 & 0.51 & 0.68 & 0.6 & 0.12 & 0.8  & 0.59 & 0.09 \\ \hline
3    & 700 & 70 & 0.7  & 0.7  & 0.7 & 0.24 & 0.81 & 0.7  & 0.18 \\ \hline
4    & 500 & 70 & 0.68 & 0.52 & 0.6 & 0.12 & 0.65 & 0.59 & 0.06 \\ \hline
\multicolumn{3}{|c|}{}     & \multicolumn{7}{c|}{Absolute value of the difference in measures}   \\ \hline
\multicolumn{3}{|c|}{$|\Psi_1-\Psi_2|$} & 0.01 & 0.18 & 0.1 & 0.06 & 0.15 & 0.09 & 0.05 \\ \hline
\multicolumn{3}{|c|}{$|\Psi_1-\Psi_3|$} & 0.2  & 0.2  & 0.2 & 0.12 & 0.16 & 0.2  & 0.09 \\ \hline
\multicolumn{3}{|c|}{$|\Psi_1-\Psi_4|$} & 0.18 & 0.02 & 0.1 & 0.06 & 0.01 & 0.09 & 0.03 \\ \hline
\multicolumn{3}{|c|}{$|\Psi_2-\Psi_3|$} & 0.19 & 0.02 & 0.1 & 0.06 & 0.01 & 0.11 & 0.04 \\ \hline
\multicolumn{3}{|c|}{$|\Psi_3-\Psi_4|$} & 0.02 & 0.18 & 0.1 & 0.06 & 0.16 & 0.11 & 0.06 \\ \hline
\end{tabular}
\end{table*}

\begin{table*}[hbt!]
\centering
\small
\caption{Values of performance metrics for four classifiers and theirs absolute value of the difference in measures -- the case of IR=0.25, P=1000, N=250.}
\label{Tab:IR25}
\begin{tabular}{|c|c|c|c|c|c|c|c|c|c|}
\hline
Meth. & TP  & TN  & HMNC       & ACC      & BACC & MCC      & $F_1$         & $G-m$  & Kappa    \\
\hline
$\Psi_1$      & 500 & 125 & 0.5  & 0.5  & 0.5  & 0    & 0.62 & 0.5  & 0    \\
$\Psi_2$      & 700 & 125 & 0.53 & 0.66 & 0.6  & 0.17 & 0.77  & 0.59 & 0.16 \\
$\Psi_3$      & 700 & 175 & 0.7 & 0.7 & 0.7 & 0.33 & 0.79  & 0.7 & 0.3 \\
$\Psi_4$      & 500 & 175 & 0.65 & 0.54 & 0.6 & 0.16 & 0.63 & 0.59 & 0.12 \\\hline
\multicolumn{3}{|c|}{}    & \multicolumn{7}{c|}{Absolute value of the difference in measures}   \\ \hline
\multicolumn{3}{|c|}{$|\Psi_1-\Psi_2|$} & 0.03 & 0.16 & 0.1  & 0.08 & 0.15 & 0.09 & 0.08 \\ \hline
\multicolumn{3}{|c|}{$|\Psi_1-\Psi_3|$}  & 0.2 & 0.2 & 0.2 & 0.16 & 0.17 & 0.2 & 0.15  \\ \hline
\multicolumn{3}{|c|}{$|\Psi_1-\Psi_4|$}  & 0.15 & 0.04 & 0.1 & 0.08 & 0.02    & 0.09 & 0.06  \\ \hline
\multicolumn{3}{|c|}{$|\Psi_2-\Psi_3|$} & 0.17 & 0.04 & 0.1 & 0.08 & 0.02    & 0.11 & 0.07 \\ \hline
\multicolumn{3}{|c|}{$|\Psi_3-\Psi_4|$} & 0.05 & 0.16 & 0.1  & 0.08 & 0.15 & 0.11  & 0.09 \\ \hline
\end{tabular}

\end{table*}

Tab.~\ref{Tab:IR01}--\ref{Tab:IR25} show a comparison of the absolute values of the quality measures for the classification of two classifiers. The classification efficiency is given as $TP$ and $TN$ values. The tables relate to three different $IR$ values ($IR=0.01$, $IR=0.1$ and $IR=0.25$). It was assumed that the majority class is a positive class. So, for example, if $IR=0.01$, $TP=500$ and $TN=5$ (see Tab.~\ref{Tab:IR01}) the method designated as $\Psi_1$  classifies $500$ out of $1000$ positive class examples correctly, as well as $5$ out of $10$ negative class examples. In each considered case of $IR$ the following relationships can be noted:

\begin{itemize}

\item  The comparison of two classifiers in which the values of $TN$ do not change. This case means that the methods differ in the accuracy in relation to the majority class classification $P$. In Tab.~\ref{Tab:IR01}--\ref{Tab:IR25}, these are the rows marked $|\Psi_1-\Psi_2|$ and $|\Psi_3-\Psi_4|$. For these rows, the difference in the absolute value of the classification quality measure is the smallest for $HMNC$  measure compared to other measures.

\item The comparison of two classifiers in which the values of $TP$ do not change. This case means that the methods differ in the accuracy in relation to the minority class classification $N$. In Tab.~\ref{Tab:IR01}--\ref{Tab:IR25}, these are the rows marked $|\Psi_1-\Psi_4|$ and $|\Psi_2-\Psi_3|$. For these rows, the difference in the absolute value of the classification quality measure is the largest for $HMNC$  measure compared to other measures.
\end{itemize}

Therefore, comparing the absolute values regarding the difference between two classification methods for different classification measures, one can indicate which of the two machine learning methods is more effective for a minority or majority class. The discussed results relate to the following values $TP>0.5*P$ and $TN>0.5*N$, which are presented in Tab.~\ref{Tab:IR01}--\ref{Tab:IR25}. Similar results apply to $TP>0.5*P$ and any $TN$ value. This results from the fact of analyzing heat maps presented in Fig.~\ref{RB:Rys_IR001}--\ref{RB:Rys_IR025}. The heat maps indicate a change in the value of the relevant measure of classification quality in the space specified by $TP$ (abscissa axis) and $TN$ (ordinate axis).

In the case of the performance measure $HMNC$, the changes are the smallest along the abscissa axis, i.e. for changing $TP$ values, which were adopted at work as a majority class. Changes in the value of measure $HMNC$ are the most significant along the ordinate axes, i.e. for changing $TN$ values that were adopted at work as a minority class. The described changes occur for $TP$ values, which are equal to approximately $\frac{TP}{P}> IR$.

In the case of values for which $\frac{TP}{P}<IR$ the heat maps for $HMNC$ metric indicate an inverse trend. This means quick changes in the value of the measure value for small changes in $TP$ and small changes in the value of the measure value for significant changes in $TN$ value.

It should also be noted that the heat map of the proposed method $HMNC$ is similar to the heat map of $G-mean$ measure. The heat map curvature (change in the direction of the isoline from horizontal to vertical) for measure $G-mean$ is independent of $IR$ factor.

In the case of the heat map for the $HMNC$ performance measure, the direction of isoline changes at $\frac{TP}{P}<IR$. The proposed measure therefore reflects $IR$ coefficients, and the analysis of the absolute values of the difference between the results of the two machine learning methods can clearly indicate which of the two methods analyzed is more effective for the minority or majority class.


Referring to the invariant properties of the classification performance metric described in~\cite{SOKOLOVA2009427}, it can be stated that:
\begin{itemize}

\item  For the correct classification of the majority classes larger than $IR$ defined by equation~(\ref{eq:IR}) the proposed $HMNC$ metric will be less sensitive to changes in the majority class and more sensitive to changes in the minority class,

\item  For the correct classification of the majority classes smaller than $IR$ defined by equation~(\ref{eq:IR}) the proposed $HMNC$ metric will be more sensitive to changes in the majority class and less sensitive to changes in the minority class.

\end{itemize}

Sensitivity to changes in classified objects is greater (in the case of the minority class) or lower (in the case of the minority class) than the change in sensitivity of other classification performance measures considered in the article, which was shown using heat maps (Fig.~\ref{RB:Rys_IR001}--\ref{RB:Rys_IR025}) and examples presented in Tab.~\ref{Tab:IR01}--\ref{Tab:IR25}.

\section{Conclusion}

In this study, we propose the new performance measure which has several properties required for the classification quality measures. The identity with other performance measures in the case of $\frac{TP}{P}=\frac{TN}{N}$ has been proven analytically. The shape of the heat map isoline indicates that it depends on $IR$ value. The proposed measure therefore has the desired properties for imbalanced data, as shown in the examples. The obtained results relate to the pairwise comparison of machine learning methods and their analysis in the context of other performance measures.

\bibliographystyle{unsrt}  
\bibliography{references}  


\end{document}